\documentclass[conference]{IEEEtran}

\IEEEoverridecommandlockouts
\usepackage{cite}
\usepackage{amsmath,amssymb}
\usepackage{booktabs}
\usepackage{array}
\newcommand{\makecell}[2][c]{\begin{tabular}{@{}#1@{}}#2\end{tabular}}
\usepackage{graphicx}
\usepackage{xspace}
\usepackage{pifont}
\usepackage{colortbl}
\usepackage{xcolor}

\newcommand{\method}{OKA-CT\xspace}
\newcommand{\ctrate}{CT-RATE\xspace}
\newcommand{\raddata}{RAD-ChestCT\xspace}

\newcommand{\auc}{AUROC\xspace}

\newcommand{\parsection}[1]{\vspace{4pt}\noindent\textbf{#1:}}

\renewcommand{\footnoterule}{%
  \kern -3pt
  \hrule width 0.35\columnwidth height 0.4pt
  \kern 2.6pt
}

\definecolor{ieeeblue}{rgb}{0.00,0.30,0.60}
\usepackage[breaklinks,colorlinks,allcolors=ieeeblue]{hyperref}
\usepackage{float}

\title{Learning Anatomy-Grounded CT Vision-Language Representations with Organ-Hierarchical Report Knowledge}

\author{
\IEEEauthorblockN{
Guoliang You\textsuperscript{1},
Hongming Li\textsuperscript{1},
Yuanwang Zhang\textsuperscript{2},
and Yong Fan\textsuperscript{1,*}
}

\IEEEauthorblockA{
\textsuperscript{1}
Department of Radiology, Perelman School of Medicine,\\
University of Pennsylvania, Philadelphia, PA 19104, USA
}

\IEEEauthorblockA{
\textsuperscript{2}
School of Engineering and Applied Science,\\
University of Pennsylvania, Philadelphia, PA 19104, USA
}
}

\begin{document}
\maketitle

\begingroup
\renewcommand{\thefootnote}{*}
\footnotetext{Corresponding author.}
\endgroup

\begin{abstract}
Medical vision-language pretraining (VLP) from paired CT images and radiology reports enables scalable representation learning, but most existing methods align either whole scans with entire reports or local image regions with text fragments. These formulations underuse a key property of radiology reports: findings are organized around anatomical structures, with abnormalities described by organs, disease concepts, locations, and severity-related attributes. We propose \method, an organ-hierarchical knowledge-augmented framework for CT-report VLP. \method first converts free-text reports into organ-conditioned knowledge using radiology report parsing and LLM-assisted semantic structuring. The extracted hierarchy is used across two learning stages. Stage~1 injects anatomy-grounded evidence into the CT visual representation through fine-grained organ-conditioned supervision, while Stage~2 uses organ-specific report evidence to guide structured report-CT contrastive learning, where hierarchy-derived semantic soft targets treat non-paired cases with shared organ-level findings as weak semantic positives rather than uniform negatives. A lightweight query-based global branch further aggregates disease-relevant volumetric evidence for whole-scan representation. On \ctrate and \raddata datasets, \method achieves zero-shot abnormality diagnosis AUROCs of 84.9 and 72.2, outperforming prior CT VLP baselines. Retrieval and patch-occlusion analyses further show improved report-image alignment and stronger sensitivity to disease-associated anatomical regions.
\end{abstract}


\begin{IEEEkeywords}
Medical Vision Language Pretraining, Organ-hierarchical Knowledge, Report Knowledge Extraction, Structured Contrastive Learning.
\end{IEEEkeywords}

\begin{figure}[t]
    \centering
    \includegraphics[width=\linewidth]{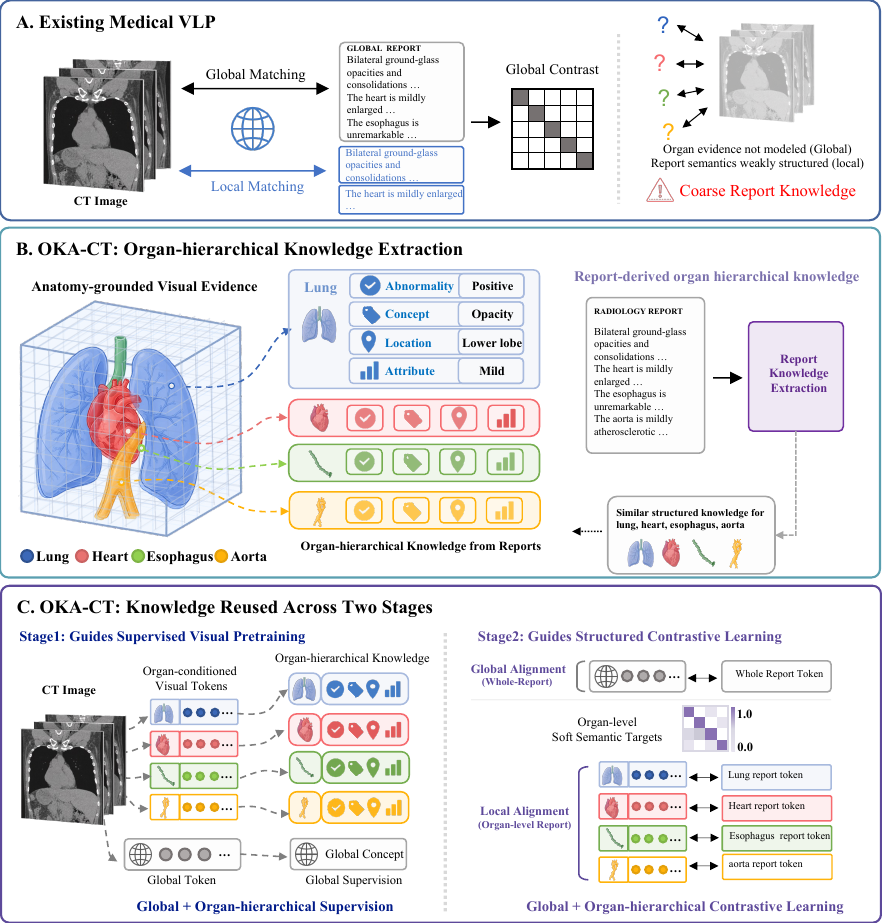}
    \vspace{-15pt}
    \caption{{From coarse report alignment to organ-hierarchical evidence learning.}
    (A) Existing CT VLP relies on global alignment or coarse local correspondence.
    (B) \method converts free-text reports into organ-hierarchical knowledge.
    (C) The hierarchy guides organ-conditioned supervision in Stage~1 and structured global and organ-level report-CT alignment in Stage~2.}
    \vspace{-12pt}
    \label{fig:teaser}
\end{figure}

\section{Introduction}
\label{sec:intro}

Medical vision-language pretraining (VLP) provides a scalable paradigm for learning clinical visual representations from paired medical images and radiology reports. This paradigm is particularly valuable for computed tomography (CT), where volumetric studies are large, findings may be sparse, and dense expert annotations are expensive to obtain. Recent CT-report VLP methods have shown that routinely collected radiology reports can support zero-shot diagnosis, image-text retrieval, and general-purpose representation learning.

Most existing medical VLP methods organize image-report supervision at either the global or local level. Global alignment methods learn a single correspondence between an entire scan and its associated report, enabling large-scale contrastive learning from paired CT-report data and supporting zero-shot diagnosis and retrieval~\cite{hamamci2025ctrate,langlotz2024merlin}. Local and fine-grained methods further align image regions, report sentences, clinical entities, or anatomical structures, showing that spatially localized evidence is important for medical image understanding ~\cite{huang2021gloria,liu2023semanticsaware,wang2025ecamp,shui2025fvlm,chen2024mg3d,zhang2025ctglip}.
However, radiology reports are not simply whole-scan descriptions or unordered collections of text fragments. They are naturally structured around anatomical systems: abnormalities are first grounded in organs, then described by disease or finding concepts, anatomical locations, severity-related attributes, and diagnostic impressions. 

This organ-hierarchical structure exposes a limitation in current global-local formulations. Whole-scan CT-report alignment compresses multi-organ findings into a single embedding and may dilute sparse but clinically important findings. Local alignment, while fine-grained, treats regions, organs, or text fragments as matching units without reusing the internal report hierarchy as a supervisory signal. As a result, two questions remain underexplored: how should organ-level report semantics shape the visual representation before language alignment, and how should contrastive learning handle non-paired cases that share anatomy-grounded disease evidence?

Our key observation is that radiology reports contain rich organ-conditioned knowledge. Report parsing can identify observations, anatomical entities, modifiers, and anatomical relations~\cite{bui2024radgraphxl}, while LLM-assisted semantic structuring can normalize free-text findings into organ-level fields. We refer to these fields as slots, including abnormality status, disease or finding concepts, anatomical location evidence, and severity-oriented attributes. These derived slots provide weak but clinically meaningful organ-conditioned knowledge, including semantic relations that connect CT images, reports, and disease labels in an anatomy-grounded evidence space. Rather than treating them as manually verified labels, we use them as structured supervision to guide visual evidence learning and to define semantic neighborhoods for contrastive alignment. 

We propose \method, an organ-hierarchical knowledge-augmented framework for CT-report VLP, as illustrated in Fig.~\ref{fig:teaser}. The central design is to extend global CT VLP with reusable organ-hierarchical knowledge. \method converts free-text reports into organ-conditioned knowledge and reuses this hierarchy across two stages. In Stage~1, fine-grained organ-conditioned visual supervision augments global disease supervision and injects anatomy-grounded evidence into the CT visual representation. In Stage~2, organ-specific report evidence guides structured report-CT contrastive learning, where hierarchy-derived semantic soft targets treat non-paired cases with shared organ-level findings as weak semantic positives rather than uniformly hard negatives. A lightweight query-based global branch further aggregates disease-relevant volumetric evidence for whole-scan representation. 

We evaluate \method on \ctrate and \raddata. Experiments show that \method improves zero-shot abnormality diagnosis, image-image retrieval, and report-image retrieval, compared with prior CT VLP baselines. Patch-occlusion analysis further shows that the learned representation is more sensitive to disease-associated anatomical regions. These results suggest that CT VLP benefits from explicitly modeling anatomy-grounded report hierarchy rather than relying only on whole-report alignment or coarse local correspondence.

Our main contributions are:
\begin{itemize}
    \item We propose \method, a CT-report VLP framework that reuses report-derived organ hierarchy for visual representation learning and report-image contrastive alignment.
    \item We construct organ-conditioned knowledge from free-text reports, including abnormality status, disease/finding concepts, anatomical location evidence, and severity-oriented attributes, via report parsing and LLM-assisted semantic structuring, without requiring manual annotation.
    \item We introduce a two-stage strategy in which organ-conditioned supervision shapes the CT visual space and hierarchy-derived soft targets define anatomy-grounded semantic neighborhoods for contrastive learning.
    \item We demonstrate improved zero-shot diagnosis and retrieval on \ctrate and \raddata, with qualitative evidence that \method is more sensitive to disease-associated anatomical regions.
\end{itemize}

\section{Related work}
\label{sec:related}

\begin{figure*}[!t]
    \centering
    \includegraphics[width=0.97\linewidth]{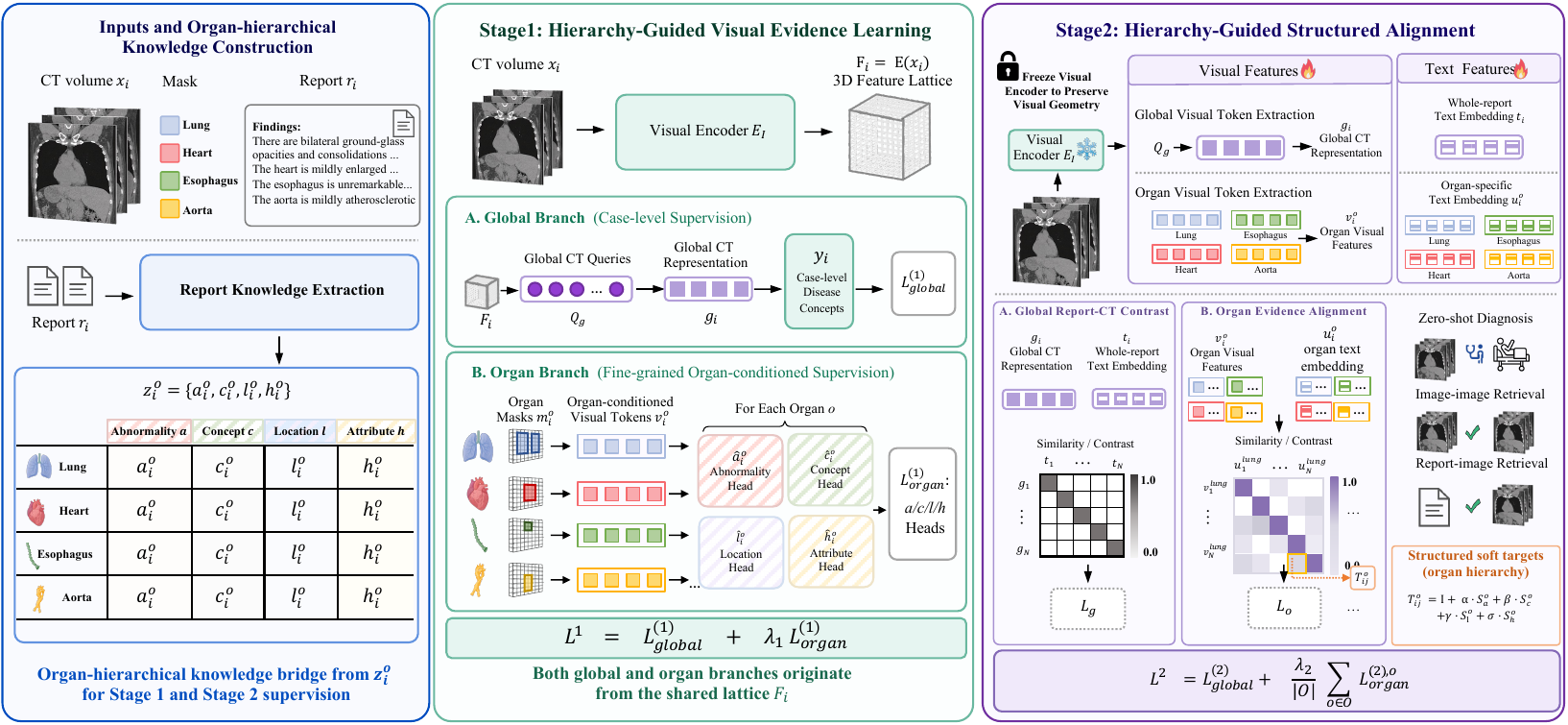}
    \vspace{-5pt}
    \caption{Overview of \method. \method extracts organ-hierarchical knowledge from reports using report parsing and LLM-assisted semantic structuring. The hierarchy contains organ abnormality status, disease/finding concepts, anatomical location evidence, and severity-oriented attributes. In Stage~1, this hierarchy augments global disease supervision by applying organ-conditioned supervision to mask-selected visual features, facilitating anatomy-grounded CT representation learning. In Stage~2, the hierarchy augments global report-CT contrastive learning by aligning organ-specific report evidence with organ visual features and constructing hierarchy-derived soft targets. The query-based global branch aggregates disease-relevant evidence for whole-volume representation.}
    \vspace{-10pt}
    \label{fig:method}
\end{figure*}

\paragraph{Global CT-report VLP}
Contrastive image-text pretraining has become a widely used paradigm for learning transferable visual representations from paired images and text ~\cite{radford2021clip,jia2021align,li2021albef}. In medical imaging, radiology reports provide scalable language supervision for learning image representations without requiring dense manual annotation, supporting diagnosis, retrieval, report generation, and downstream transfer across chest radiography and other medical imaging settings~\cite{wang2017chestxray8,irvin2019chexpert,johnson2019mimiccxr,demnerfushman2015openi,sun2022medclip}. 
CT-report VLP is particularly challenging because CT volumes are three-dimensional, computationally expensive, and often contain sparse findings distributed across multiple anatomical structures. Recent CT VLP methods have demonstrated the value of large-scale whole-volume report alignment. CT-RATE introduced a large-scale paired CT-report resource and demonstrated that contrastive learning from 3D CT volumes and reports can support zero-shot diagnosis and retrieval~\cite{hamamci2025ctrate}; Merlin studied 3D CT representation learning using radiology reports and structured clinical supervision~\cite{langlotz2024merlin}.
Other methods improve CT representation learning and understanding through cross-modal knowledge distillation, semantic alignment, and scalable radiology supervision~\cite{cao2024biud,lai2025brgsa}.

\paragraph{Fine-grained, Local, Anatomy-Aware Medical VLP}
To address the limitations of global image-report alignment, many medical VLP methods introduce finer-grained correspondence between visual regions and textual evidence. Global-local methods align image patches or regions with report words, phrases, or sentences, improving label-efficient recognition and localized representation learning~\cite{huang2021gloria,muller2022lovt,boecking2022making,wang2025ecamp}. For CT, fVLM aligns anatomical regions with corresponding report descriptions~\cite{shui2025fvlm}, MG-3D models multi-grained report semantics and inter-patient relationships~\cite{chen2024mg3d}, and CT-GLIP constructs grounded organ-level CT-report pairs for full-body scenarios ~\cite{zhang2025ctglip}. These studies demonstrate the value of local anatomical correspondence for medical image understanding~\cite{huang2024bcnet,gong2025boundary}.

\paragraph{Structured Organ Knowledge and Semantic Contrastive Learning}
Radiology reports contain structured clinical information, including observations, anatomical entities, modifiers, uncertainty, and negation. Report parsing methods extract entities and relations from free-text reports~\cite{bui2024radgraphxl}, and structured text semantics improve biomedical VLP~\cite{boecking2022making,wang2025ecamp}.
Standard image-text contrastive learning treats only the paired image-report sample as positive and all others as negative. In medical VLP, this may introduce false-negative supervision because patients may share the same abnormality, organ involvement, or anatomical pattern. Semantics-aware VLP methods address this issue by incorporating label- or text-derived relationships among samples~\cite{liu2023semanticsaware,sun2022medclip}.

\section{Method}
\label{sec:method}

\subsection{Overview}
\method learns an anatomy-grounded evidence space for CT-report VLP. Given a CT volume $x_i$, its paired report $r_i$, case-level disease labels $y_i$, and organ masks $\{m_i^o\}_{o\in\mathcal{O}}$, the model produces a global CT representation $g_i$ and organ-conditioned visual tokens $\{v_i^o\}_{o\in\mathcal{O}}$. In our chest CT experiments, the organ set $\mathcal{O}$ contains lung, heart, esophagus, and aorta. 
The central component of \method is report-derived organ-hierarchical knowledge.  A report knowledge extraction pipeline converts each report $r_i$ into organ-conditioned structured knowledge ${z_i^o=\{a_i^o,c_i^o,\ell_i^o,h_i^o\}}$, where ${a_i^o}$ denotes organ abnormality status, ${c_i^o}$ denotes disease/finding concepts, ${\ell_i^o}$ denotes anatomical location evidence, and ${h_i^o}$ denotes severity-oriented attributes.
\method uses this knowledge hierarchy to augment two complementary learning stages: Stage~1 uses the hierarchy to enhance organ-conditioned visual evidence learning, while Stage~2 uses it to complement paired global report-CT contrastive learning with organ evidence alignment and batch-level semantic soft targets, as illustrated in Fig.~\ref{fig:method}.

\subsection{Report Parsing and LLM-assisted Semantic Structuring}
\begin{figure}[t]
    \centering
    \includegraphics[width=\linewidth]{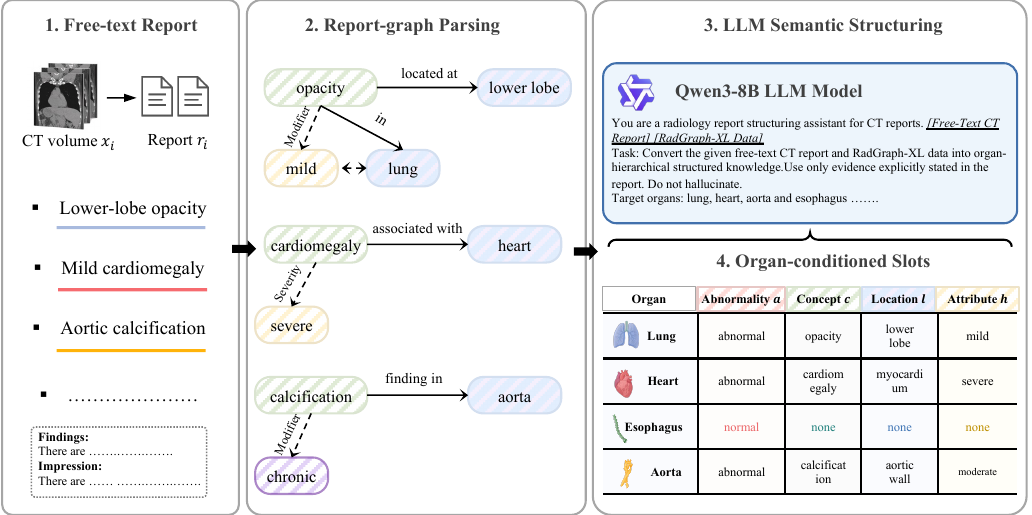}
    \vspace{-12pt}
    \caption{Report parsing and semantic structuring. A report parser first extracts clinical entities and relations from free-text reports, and an LLM-assisted semantic structuring step then normalizes them into organ-conditioned fields: abnormality, disease/finding concept, anatomical location, and severity-oriented attributes. The resulting organ hierarchy is reused for Stage~1 organ-conditioned visual supervision and Stage~2 organ evidence alignment with hierarchy-derived semantic soft targets.}
    \vspace{-10pt}
    \label{fig:rg_qwen3_knowledge}
\end{figure}

The report knowledge extraction pipeline converts each free-text radiology report into organ-conditioned structured knowledge. Given report $r_i$ and target organ set $\mathcal{O}$, we first apply a radiology report parser to extract clinical entities and relations, including observations, anatomical entities, modifiers, and uncertainty cues~\cite{bui2024radgraphxl}. 
We then use an LLM-assisted semantic structuring step to map the extracted information and original report text into the organ-level schema. As illustrated in Fig.~\ref{fig:rg_qwen3_knowledge}, the knowledge extraction pipeline produces an organ-conditioned schema for each organ $o$: $z_i^o=\{a_i^o,c_i^o,\ell_i^o,h_i^o\}$, where the four slots represent abnormality status, disease or finding concepts, anatomical location evidence, and severity-oriented modifier evidence, respectively. When organ-specific concepts are missing or ambiguous, case-level disease labels are used as conservative fallback supervision when applicable.

These slots are automatically derived structured knowledge rather than manually verified organ-level ground truth. Therefore, \method does not treat them as exhaustive dense labels for every organ and every case. Instead, the hierarchy is used to supervise organ-conditioned visual representation learning in Stage~1 and to define soft semantic relations among organ-level samples for structured contrastive learning in Stage~2.

\subsection{Stage~1: Hierarchy-guided Visual Evidence Learning}
backboneStage~1 trains a CT visual encoder to encode both global disease evidence and organ-conditioned anatomical evidence. The image encoder $E_I$, instantiated as a 3D ResNet-18 (R3D-18) backbone~\cite{3dresnet}, maps a CT volume ${x_i}$ to a 3D feature map $F_i=E_I(x_i)$. A query-based global branch aggregates the 3D feature maps into a global representation $g_i$. A global prediction head $\phi_g$ maps $g_i$ to case-level disease logits, supervised by disease concept labels $y_i$:
\begin{equation}
    \mathcal{L}_{\mathrm{global}}^{(1)}=
    \ell_{\mathrm{BCE}}\!\left(\phi_g(g_i),y_i\right).
\end{equation}
This global branch provides scan-level disease separability.

To inject organ-hierarchical evidence, \method uses organ masks to associate report-derived organ knowledge with the anatomical regions in the CT feature lattice. Each organ mask ${m_i^o}$ is downsampled to the resolution of ${F_i}$. For organ $o$, we first select the features inside the organ region and then aggregate them into an organ-conditioned visual representation:
\begin{equation}
    v_i^o=Pool(\{F_i(p)\mid m_i^o(p)=1\}),
\end{equation}
where $p$ indexes spatial locations in the feature lattice and ${Pool(\cdot)}$ denotes masked feature aggregation. Organ supervision is applied only when the downsampled mask contains sufficient foreground support, indicated by \(\mathbf{1}_{i,o}\) .

For each valid organ representation \(v_i^o\), organ-specific prediction heads estimate the four hierarchy fields of \(z_i^o=\{a_i^o,c_i^o,\ell_i^o,h_i^o\}\):
\begin{equation}
    \hat a_i^o=\psi_a^o(v_i^o),\quad
    \hat c_i^o=\psi_c^o(v_i^o),\quad
    \hat \ell_i^o=\psi_\ell^o(v_i^o),\quad
    \hat h_i^o=\psi_h^o(v_i^o),
\end{equation}
where \(\psi_a^o,\psi_c^o,\psi_\ell^o,\psi_h^o\) denote the organ-specific prediction heads. The organ-level supervision is defined as
\begin{equation}
\begin{split}
\mathcal{L}_{\mathrm{organ}}^{(1)}
=\sum_{o\in\mathcal{O}}\mathbf{1}_{i,o}\big[
&\mathcal{L}_{\mathrm{BCE}}(\hat a_i^o,a_i^o)
+\mathcal{L}_{\mathrm{BCE}}(\hat c_i^o,c_i^o)\\
&+\mathcal{L}_{\mathrm{CE}}(\hat \ell_i^o,\ell_i^o)
+\mathcal{L}_{\mathrm{CE}}(\hat h_i^o,h_i^o)
\big],
\end{split}
\end{equation}
where \(\mathbf{1}_{i,o}\) indicates whether organ \(o\) has sufficient foreground support, with binary cross-entropy for prediction of abnormality status and disease/finding concept, and cross-entropy for prediction of anatomical location evidence and severity-oriented attribute when categorical labels are available. The full Stage~1 objective is: 
\begin{equation}
    \mathcal{L}^{(1)}=
    \mathcal{L}_{\mathrm{global}}^{(1)}
    +\lambda_1\mathcal{L}_{\mathrm{organ}}^{(1)}.
\end{equation}
By combining scan-level disease supervision with organ-conditioned hierarchy supervision, Stage~1 encourages the visual encoder to learn anatomy-conditioned visual representations that are both globally discriminative and anatomically grounded at the organ-level.


\begin{table*}[!t]
\caption{Zero-shot abnormality diagnosis. Published baseline numbers are collected from prior CT VLP comparisons. \method denotes our full organ-hierarchical framework. $^{\dagger}$COLIPRI-C denotes the contrastive-only COLIPRI variant.}
\label{tab:diagnosis_results}
\centering
\footnotesize
\setlength{\tabcolsep}{17pt}
\renewcommand{\arraystretch}{1.15}
\begin{tabular}{l|c|c|c|c|c|c|c|c}
\toprule
Method & \multicolumn{4}{c|}{CT-RATE$\uparrow$} &
\multicolumn{4}{c}{RAD-ChestCT$\uparrow$} \\
\cline{2-9}
 & \auc & ACC & F1 & Prec. & \auc & ACC & F1 & Prec. \\
\hline
CT-Net~\cite{draelos2021radchestct} & 60.3 & 58.1 & 63.1 & 23.9 & 54.4 & 54.0 & 58.7 & 28.5 \\
\hline
CT-CLIP~\cite{hamamci2025ctrate} & 73.1 & 66.8 & 70.7 & 32.3 & 62.9 & 59.5 & 64.2 & 33.6 \\
\hline
BIUD~\cite{cao2024biud} & 71.3 & 68.1 & 71.6 & 33.8 & 62.9 & 60.6 & 65.2 & 33.7 \\
\hline
Merlin~\cite{langlotz2024merlin} & 72.8 & 67.2 & 70.9 & 33.7 & 64.4 & 61.9 & 66.3 & 34.8 \\
\hline
fVLM~\cite{shui2025fvlm} & 77.8 & 71.8 & 75.1 & 37.9 & 68.0 & 64.7 & 68.8 & 37.4 \\
\hline
COLIPRI-C$^{\dagger}$~\cite{wald2025colipri} & 76.3 & -- & -- & -- & 69.1 & -- & -- & -- \\
\hline
\rowcolor[gray]{.92}\textbf{\method (Ours)} & \textbf{84.9} & \textbf{78.2} & \textbf{80.5} & \textbf{44.6} & \textbf{72.2} & \textbf{67.1} & \textbf{70.8} & \textbf{39.0} \\
\bottomrule
\end{tabular}
\vspace{-10pt}
\end{table*}

\subsection{Stage~2: Hierarchy-guided Structured Alignment}

\begin{figure}[t]
    \centering
    \includegraphics[width=\linewidth]{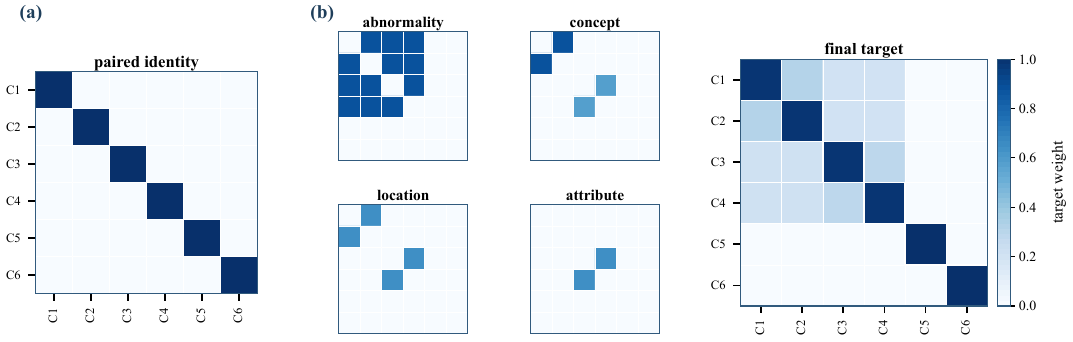}
    \vspace{-12pt}
    \caption{Organ-hierarchical soft target. For each organ, report-derived abnormality, disease/finding concept, location, and severity-oriented attribute slots define same-organ semantic neighborhoods. The paired diagonal remains strongest, while off-diagonal entries connect non-paired samples that share anatomy-grounded evidence, turning one-hot matching into paired identity plus hierarchy-defined semantic neighborhood learning.}
    \vspace{-10pt}
    \label{fig:structured_target_matrix}
\end{figure}
Stage~2 learns report-CT alignment at both the global and organ levels while preserving the anatomy-aware visual geometry acquired in Stage~1. We freeze the visual encoder and train the projection and text-alignment components for global and organ-level contrastive learning, allowing report alignment to be learned on top of this structured visual space rather than overwriting it.

At the global level, the text encoder maps each report ${r_i}$ to a report embedding ${t_i}$, and the global CT-report similarity between CT volume $i$ and report $j$ is
\begin{equation}
    s^g_{ij}=\tau_g^{-1}\langle g_i, t_j\rangle,
\end{equation}
where \(\tau_g\) is a temperature parameter. 
The global contrastive loss follows the bidirectional image-text objective:
\begin{equation}
\mathcal{L}_{global}^{(2)}
=
-\frac{1}{2B}\sum_{i=1}^{B}
\bigg[
\log \tfrac{\exp(s^g_{ii})}{\sum_{j=1}^{B}\exp(s^g_{ij})}
+
\log \tfrac{\exp(s^g_{ii})}{\sum_{j=1}^{B}\exp(s^g_{ji})}
\bigg].
\end{equation}

At the organ level, the text encoder also produces an organ-specific text embedding ${u_i^o}$ from the textualized evidence of organ o. The organ-level similarity between CT organ representation ${v_i^o}$ 
and organ-specific report embedding ${u_j^o}$ is
\begin{equation}
    s^o_{ij}=\tau_o^{-1}\langle \bar v_i^o,\bar u_j^o\rangle ,
\end{equation}
where \(\bar v_i^o\) and \(\bar u_j^o\) are \(\ell_2\)-normalized organ visual and text embeddings, respectively, and \(\tau_o\) is a temperature parameter.

Instead of treating every non-paired sample as a hard negative, \method uses the report-derived hierarchy to construct an organ-specific soft target matrix. 
For each hierarchy field ${f\in\{a,c,\ell,h\}}$, we define a same-organ relation matrix:
\begin{equation}
    S_f^o(i,j)=\mathbf{1}\!\left[
    \operatorname{overlap}\left(f_i^o,f_j^o\right)>0
    \right],
\end{equation}
where $f_i^o$ denotes the value of field $f$ for organ $o$ in case $i$.
A positive entry indicates that two cases share semantic evidence for the same organ and hierarchy field.

The raw organ-specific target is:
\begin{equation}
\label{eq:target}
    T^o=I
    +\alpha S_a^o
    +\beta S_c^o
    +\gamma S_\ell^o
    +\sigma S_h^o .
\end{equation}
where \(S_a^o\), \(S_c^o\), \(S_\ell^o\), and \(S_h^o\) denote abnormality status, disease/finding concept, anatomical location evidence, and severity-oriented attribute relations, respectively. 
The identity matrix ${I}$ preserves the paired CT-report sample as the strongest positive, while off-diagonal entries assign weak positive weights to non-paired samples with shared anatomy-grounded organ-level evidence.
In our implementation, abnormality status and disease/finding concept overlap receive higher weights because they carry stronger diagnostic semantics, whereas location and severity-oriented attribute overlap serve as conservative refinements. 

Given a logit matrix $S$ and a non-negative target matrix $A$, we define the weighted row-wise contrastive loss to incorporate hierarchy-derived soft targets for organ-level alignment:
\begin{equation}
    \mathcal{H}(S,A)=
    -\frac{1}{n}
    \sum_{i=1}^{n}\sum_{j=1}^{n}A_{ij}
    \log
    \frac{\exp(S_{ij})}{\sum_{k=1}^{n}\exp(S_{ik})},
\end{equation}
In particular, this loss uses the raw target mass rather than row-normalized targets, preserving the number of reliable semantic neighbors for each sample.

For each organ \(o\), the structured contrastive loss is defined with the hierarchy-derived target:
\begin{equation}
\label{eq:softloss}
    \mathcal{L}_{\mathrm{organ}}^{(2),o}
    =\frac{1}{2}\left[
    \mathcal{H}(s^o,T^o)+
    \mathcal{H}((s^o)^\top,(T^o)^\top)
    \right].
\end{equation}
The full Stage~2 objective is:
\begin{equation}
    \mathcal{L}^{(2)}=
    \mathcal{L}_{\mathrm{global}}^{(2)}
    +\lambda_2\frac{1}{|\mathcal{O}|}
    \sum_{o\in\mathcal{O}}\mathcal{L}_{\mathrm{organ}}^{(2),o}.
\end{equation}
This objective extends paired report-CT contrastive learning by incorporating anatomy-grounded semantic neighborhoods. Cases with shared organ-level findings are no longer forced to be uniformly negative, which better reflects the clinical structure of radiology reports.

\subsection{Query-based Global Branch}

The global visual representation should capture disease-relevant volumetric evidence rather than rely only on spatial average pooling over the 3D feature map. \method therefore uses a lightweight query-based global branch to summarize the shared spatial feature map $F_i$. Given the 3D feature lattice $F_i$, a set of learnable global queries attends to the spatial feature grid and the attended query features are pooled and projected to form the global representation:
\begin{equation}
    q_i=\mathrm{Attn}(Q_g,F_i),\qquad
    g_i=W_g\,\mathrm{Pool}(q_i).
\end{equation}
Here, \(Q_g\) denotes learnable global queries and \(W_g\) projects the pooled query features into the global embedding space. 
The resulting \(g_i\) is used for case-level disease classification in Stage~1 and global CT-report contrastive learning in Stage~2. 
This branch complements organ-conditioned supervision by allowing the model to aggregate whole-volume cues that may span multiple anatomical regions.

\section{Experiments}
\label{sec:experiments}

\subsection{Datasets and Metrics}
\parsection{Datasets}
We evaluate \method on two public chest CT datasets: \ctrate and \raddata. \ctrate is a large-scale chest CT dataset containing 25,692 non-contrast 3D scans from 21,304 patients, expanded to 50,188 reconstructed volumes with paired radiology reports and abnormality labels~\cite{hamamci2025ctrate}. Following CT VLP evaluation protocols, we use \ctrate for zero-shot abnormality diagnosis and image/report retrieval.  
We further evaluate cross-dataset generalization on \raddata, a non-contrast chest CT abnormality dataset from Duke University, comprising 36,316 volumes collected between 2012 and 2017 with 83 abnormality labels ~\cite{draelos2021radchestct}. Following prior CT VLP evaluation protocols~\cite{hamamci2025ctrate}, we report results on the public subset of 3,630 scans. 

\parsection{Evaluation Tasks}
We consider three complementary tasks. Zero-shot abnormality diagnosis scores each CT against disease prompts and evaluates whether the learned image representation separates disease axes without task-specific finetuning. Image-image retrieval ranks CT candidates for each CT query, where disease-label overlap defines relevance. Report-image retrieval ranks CT candidates for each report query and evaluates whether the paired CT is retrieved.

\parsection{Metrics}
For zero-shot diagnosis, we report macro AUROC over the 18 CT-RATE abnormality categories, together with accuracy (ACC), F1 score, and precision when following the published comparison protocol. For image-image retrieval, we report MAP@5/10/50; for report-image retrieval, we report Recall@5/10/50/100. We additionally conduct patch-occlusion analysis to assess whether disease scores are sensitive to anatomy-relevant regions.

\subsection{Implementation details}
The CT image encoder is instantiated as a 3D ResNet-18 CT backbone and trained using the two-stage procedure described in Sec.~\ref{sec:method}. Automatic organ masks are generated using a CT segmentation pipeline~\cite{wasserthal2023totalsegmentator} and grouped into four chest organs: lung, heart, esophagus, and aorta. These masks are used to extract organ-conditioned visual representations from the shared 3D feature lattice.
The report knowledge extraction pipeline uses RadGraph-XL~\cite{bui2024radgraphxl} for radiology entity and relation extraction and Qwen3-8B~\cite{yang2025qwen3} for LLM-assisted semantic structuring.
For each target organ, the pipeline converts free-text reports into organ-conditioned fields, including abnormality status, disease/finding concepts, anatomical location evidence, and severity-oriented attributes. 
In preliminary comparisons across model scales, Qwen3-8B offered a balanced trade-off between extraction quality and computational cost. \ctrate disease labels are used as global disease supervision and as conservative fallback information when report-derived organ concepts are missing.

In Stage~1, the visual encoder is trained with both global case-level disease supervision and organ-conditioned supervision derived from the extracted hierarchy. In Stage~2, the visual encoder is frozen to preserve the anatomy-aware visual geometry learned in Stage~1, while global and organ-level report-CT contrastive alignment is optimized. Unless otherwise stated, we set the Stage~1 organ-supervision weight to $\lambda_1$=0.05 and the Stage~2 organ-alignment weight to $\lambda_2$=0.1, $\alpha=0.8, \beta=0.8, \gamma=0.2, \sigma=0.15$.

\begin{table}[t]
\caption{CT-RATE retrieval. VocabFine and ClassFine are retrieval variants reported with the CT-CLIP work. Other published baseline numbers follow prior CT VLP comparisons.}
\label{tab:retrieval_results}
\centering
\footnotesize
\setlength{\tabcolsep}{2.4pt}
\renewcommand{\arraystretch}{1.15}
\begin{tabular}{p{0.31\linewidth}|p{0.30\linewidth}|p{0.33\linewidth}}
\toprule
Method &
\makecell{Img2Img MAP\\@5/10/50$\uparrow$} &
\makecell{Rpt2Img Recall\\@5/10/50/100$\uparrow$} \\
\hline
CT-Net~\cite{draelos2021radchestct} & 59.4/48.1/40.7 & --/--/--/-- \\
\hline
VocabFine~\cite{hamamci2025ctrate} & 68.3/57.2/48.8 & 0.1/0.6/2.3/2.0 \\
\hline
ClassFine~\cite{hamamci2025ctrate} & 67.9/56.8/48.5 & --/--/--/-- \\
\hline
CT-CLIP~\cite{hamamci2025ctrate} & 68.3/57.2/48.9 & 2.9/5.0/18.0/28.7 \\
\hline
Merlin~\cite{langlotz2024merlin} & 62.6/51.3/43.9 & 1.5/2.7/7.7/12.7 \\
\hline
\rowcolor[gray]{.92}\textbf{\method (Ours)} & \textbf{71.1/61.6/54.1} & \textbf{11.9/18.6/42.9/57.1} \\
\bottomrule
\end{tabular}
\vspace{-5pt}
\end{table}

\subsection{Main results}

\parsection{Zero-shot Abnormality Diagnosis}
Table~\ref{tab:diagnosis_results} compares \method with prior CT VLP baselines on CT-RATE and RAD-ChestCT. On CT-RATE, \method achieves a macro AUROC of 84.9, ACC of 78.2, F1 score of 80.5, and precision of 44.6, outperforming the listed complete-metric CT VLP baselines across all four diagnosis metrics and improving macro AUROC by 7.1 percentage points over fVLM~\cite{shui2025fvlm} and by 8.6 percentage points over the contrastive-only COLIPRI variant ~\cite{wald2025colipri}. On RAD-ChestCT, \method obtains a macro AUROC of 72.2, ACC of 67.1, F1 score of 70.8, and precision of 39.0. These results indicate that the learned organ-hierarchical representation transfers beyond the CT-RATE training distribution. The cross-dataset improvement is smaller than the CT-RATE gain, suggesting that domain shift, label-space differences, and calibration remain important directions for future improvement. Overall, the diagnosis results show that organ-hierarchical knowledge is not merely an auxiliary local signal. Instead, the anatomy-grounded supervision introduced in \method improves the global CT representation used for zero-shot disease recognition.

\begin{table}[t]
\caption{Component ablations on CT-RATE. S1 denotes Stage~1 visual supervision and S2 denotes Stage~2 report-CT contrastive alignment. 
We report the diagnosis and retrieval metrics.}
\label{tab:component_ablation}
\centering
\footnotesize
\setlength{\tabcolsep}{8.5pt}
\renewcommand{\arraystretch}{1.15}
\begin{tabular}{l|c|c|c}
\toprule
Variant 
& \makecell{Diag.\\macro \auc$\uparrow$}
& \makecell{Img2Img\\MAP@50$\uparrow$}
& \makecell{Rpt2Img\\R@100$\uparrow$} \\
\hline
S2 contrast only 
& 76.8 & 47.1 & 32.0 \\
\hline
+ S1 w/ concept 
& 80.7 & \textbf{54.2} & 55.0 \\
\hline
+ S1 w/ hierarchy 
& 83.6 & \textbf{54.2} & 52.9 \\
\hline
\rowcolor[gray]{.92}
\textbf{Full \method} 
& \textbf{84.9} & 54.1 & \textbf{57.1} \\
\bottomrule
\end{tabular}
\end{table}

\parsection{Retrieval Performance}
Table~\ref{tab:retrieval_results} reports CT-RATE retrieval performance. For image-image retrieval, \method achieves MAP@5, MAP@10, and MAP@50 scores of 71.1, 61.6, and 54.1, respectively. These results indicate that \method learns a CT embedding space with stronger disease-level neighborhood structure than prior baselines. For report-image retrieval,  \method achieves Recall@5, Recall@10, Recall@50, and Recall@100 scores of 11.9, 18.6, 42.9, and 57.1, respectively. The improvement is particularly large for report-image retrieval, which is more sensitive to whether textual findings can be correctly associated with visual evidence. This supports the central motivation of \method: report-derived organ hierarchy improves not only disease classification, but also cross-modal alignment between reports and CT volumes.

\subsection{Ablation and Analysis}

\begin{figure*}[t]
    \centering
    \includegraphics[width=\linewidth]{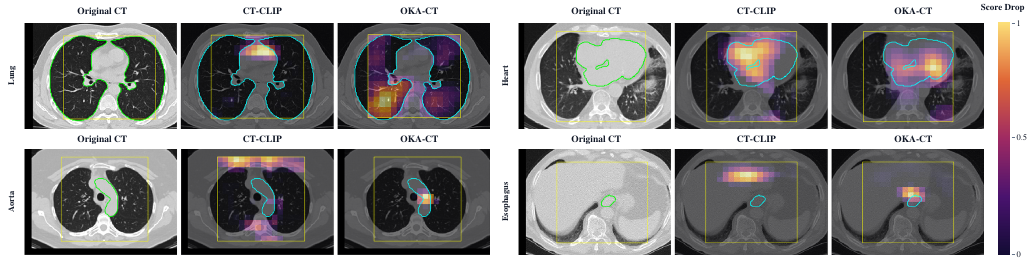}
    \vspace{-5pt}
    \caption{Patch-occlusion visualization. Each organ block shows a representative case and compares the original CT slice, CT-CLIP, and \method. Warmer colors indicate larger positive disease-score drops after local patch occlusion. The yellow box marks the model-visible crop; heatmaps are only defined inside this crop. \method shows stronger responses around disease-associated organ regions.}
    \vspace{-10pt}
    \label{fig:occlusion_gallery}
\end{figure*}

\begin{table}[t]
\caption{Stage~1 visual representation. CT-RATE diagnosis performance before Stage~2 CT-report contrastive alignment. }
\label{tab:stage1_visual_checkpoint}
\centering
\footnotesize
\setlength{\tabcolsep}{8.5pt}
\renewcommand{\arraystretch}{1.15}
\begin{tabular}{l|c}
\toprule
Stage~1 supervision & \makecell{Diag.\\macro \auc$\uparrow$} \\
\hline
Global concept supervision only & 86.7 \\
\hline
Global concept + organ-hierarchical supervision & 87.5 \\
\bottomrule
\end{tabular}
\vspace{-10pt}
\end{table}

\begin{figure}[b]
    \centering
    \includegraphics[width=0.85\linewidth]{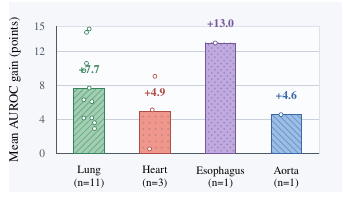}
    \vspace{-5pt}
    \caption{Organ-group deltas. Bars report the mean per-disease AUROC improvement over Stage~2-only training within each anatomical group, with $n$ denoting the number of assigned disease concepts. Open circles denote individual disease-level deltas.}
    \vspace{-5pt}
    \label{fig:organ_delta}
\end{figure}

\parsection{Stage~1 Hierarchy Supervision}
Table~\ref{tab:stage1_visual_checkpoint} evaluates the visual representation after Stage~1, before Stage~2 CT-report contrastive alignment. The global-only variant, trained with case-level disease concept supervision, achieves a macro AUROC of 86.7 using the classification head. Adding organ-hierarchical supervision improves this Stage~1 AUROC to 87.5. This comparison shows that report-derived organ hierarchy improves the CT visual representation. 
It is worth noting that Table~\ref{tab:stage1_visual_checkpoint} evaluates the supervised Stage~1 visual classifier and is therefore not directly comparable to the zero-shot image-text diagnosis results in Table~\ref{tab:diagnosis_results}. Table~\ref{tab:diagnosis_results} evaluates zero-shot VLP performance after report-CT alignment, whereas Table~\ref{tab:stage1_visual_checkpoint} isolates the quality of the supervised visual representation before Stage~2.

\parsection{Component Ablations}
Table~\ref{tab:component_ablation} reports a stepwise ablation of the two-stage design on CT-RATE. 
Stage~2 contrastive learning alone achieves a diagnosis macro AUROC of 76.8, image-image MAP@50 of 47.1, and report-image Recall@100 of 32.0. Adding Stage~1 concept supervision improves macro AUROC to 80.7 and report-image Recall@100 to 55.0, confirming the importance of supervised visual representation learning. Replacing concept-only Stage~1 supervision with organ-hierarchical supervision further improves macro AUROC to 83.6, indicating that abnormality status, disease/finding concepts, location evidence, and severity-oriented attributes provide useful anatomy-grounded visual supervision. The full \method model, which additionally uses hierarchy-derived soft targets in Stage~2, achieves the best macro AUROC of 84.9 and the best report-image Recall@100 of 57.1. These results show that the organ hierarchy contributes at both stages: first by shaping the CT visual space, and then by improving structured report-CT contrastive alignment.

\parsection{Patch-occlusion evidence}
We further conduct patch-occlusion analysis to assess whether \method relies on anatomy-relevant visual evidence. The analysis is performed on full 3D CT volumes for representative cases covering 16 disease concepts mapped to one of four organs: lung, heart, esophagus, and aorta. For visualization, we display the axial slice with the largest target-organ mask, while the model input remains the full 3D CT volume. For each disease, we define the disease score as the image-text similarity to the positive disease prompt minus the similarity to the corresponding negative prompt. We then occlude local \(24{\times}24\) patches with a stride of 12 on a three-slice slab centered at the display slice and measure the score drop. Positive score drops indicate regions that contribute to the disease prediction and are projected back to the display slice to form an occlusion heatmap. As shown in Fig.~\ref{fig:occlusion_gallery}, \method produces stronger score-drop responses around disease-associated organ regions than CT-CLIP. This suggests that organ-hierarchical supervision encourages the global CT representation to rely more heavily on anatomy-aligned visual evidence.

\parsection{Organ-group Behavior}
Fig.~\ref{fig:organ_delta} reports per-disease AUROC improvements grouped by anatomical target organ, using Stage~2-only training as the reference. \method improves average AUROC for lung, heart, esophagus, and aorta groups by 7.7, 4.9, 13.0, and 4.6 points, respectively. Although the esophagus and aorta groups contain fewer disease concepts, the improvements across all four organ groups suggest that the gains of \method are not confined to a single anatomical structure. This supports the proposed mechanism that the organ-hierarchical supervision reshapes clinically meaningful evidence axes across the chest CT volume and improves disease recognition across multiple anatomical systems.
\section{Discussion and Limitations}
\label{sec:discussion}

The results support our motivation that organ-hierarchical knowledge provides a supervision signal for CT-report vision-language learning. In \method, this hierarchy is reused across stages: it guides anatomy-conditioned visual representation learning in Stage~1 and defines organ-level semantic relationships for structured contrastive learning in Stage~2, where clinically similar non-paired cases can be treated as weak positives rather than uniform negatives. This design improves organ-level alignment and whole-volume diagnosis. The structured slots are derived from radiology reports, offering a scalable and clinically meaningful source of organ-level supervision without requiring dense manual annotation. Meanwhile, they are used as weak semantic signals rather than curated expert labels, and their granularity, calibration, and normalization can be further improved. 
Our current study is also limited to CT-RATE and four chest organs, i.e., lung, heart, esophagus, and aorta; future work will extend the organ vocabulary and evaluate broader datasets and CT protocols, including contrast-enhanced, multi-phase, and whole-body CT. This collective use of slots across organs and hierarchy fields also reduces reliance on any single extracted label.

\section{Conclusion}
\label{sec:conclusion}

Overall, \method demonstrates that the organ-hierarchical report knowledge can serve as weak supervision for CT-report VLP. By injecting organ-conditioned knowledge derived from free-text reports into visual representation learning and using the hierarchy to guide structured contrastive report-image alignment,  \method improves zero-shot diagnosis and retrieval while showing stronger sensitivity to disease-associated organ regions. These results suggest that CT VLP benefits from anatomy-grounded semantic evidence in addition to whole-volume report matching. The main limitations lie in the quality of automatically extracted report knowledge, the restricted anatomical vocabulary, and the need for broader cross-institutional validation. Addressing these limitations may strengthen organ-hierarchical learning for scalable and clinically reliable CT vision-language pretraining.

{
    \bibliographystyle{IEEEtran}
    \bibliography{main}
}
\end{document}